\documentclass[10pt, conference]{ieeeconf}
\IEEEoverridecommandlockouts
\usepackage{booktabs}
\usepackage{graphicx} 
\usepackage{amsmath}
\usepackage{amssymb}
\usepackage{xcolor}
\usepackage{hyperref}
\usepackage{tabularx}
\usepackage{pifont}
\usepackage{siunitx}
\newcommand{\cmark}{\ding{51}}%
\newcommand{\xmark}{\ding{55}}%
\newcommand{\rcs}{RCS}
\newcommand{\vla}{VLA}

\newcommand{\pickboxtask}{Pick-Cuboid}
\newcommand{\picktoytask}{Pick-Toy}
\newcommand{\franka}{FR3}
\newcommand{\xarm}{xArm7}
\newcommand{\ur}{UR5e}
\newcommand{\so}{SO101}
\newcommand{\pizero}{\texorpdfstring{$\pi_0$}{Pi zero}}
\newcommand{\ovla}{OpenVLA}
\newcommand{\octo}{Octo}

\title{\LARGE \bf
Robot Control Stack: A Lean Ecosystem for Robot Learning at Scale
}

\author{Tobias J\"ulg$^{*1}$, Pierre Krack$^{*1}$, Seongjin Bien$^{*1}$, Yannik Blei$^{1}$, Khaled Gamal$^{1}$, Ken Nakahara$^{2}$,\\ Johannes Hechtl$^{1,3}$, Roberto Calandra$^{2}$, Wolfram Burgard$^{1}$ and Florian Walter$^{1,4}$
\thanks{$^*$Equal contribution. $^{1}$Department of Computer Science~\& Artificial Intelligence, University of Technology Nuremberg, Germany. Contact: \texttt{tobias.juelg@utn.de}. $^{2}$Learning, Adaptive Systems and Robotics (LASR) Lab, Faculty of Computer Science, TU Dresden, Germany. $^{3}$Siemens Foundational Technologies, Siemens AG, Germany. $^{4}$Chair for Robotics, Artificial Intelligence and Real-Time Systems, TUM School of Computation, Information and Technology, Technical University of Munich, Germany. 
This work has been partially supported by the projects GeniusRobot (grant no.~01IS24083) and the Robotics Institute Germany (RIG) (grant no.~16ME1006) funded by the German Federal Ministry of Research, Technology and Space (BMFTR).
The authors acknowledge the HPC resources provided by the Erlangen National HPC Center (NHR@FAU) under the BayernKI project no.~v106be. We thank Suman Navaratnarajah for his code contributions to RCS.}
}

\begin{document}

\maketitle
\thispagestyle{empty}
\pagestyle{empty}

\begin{abstract}
Vision-Language-Action models (\vla{}s) mark a major shift in robot learning.
They replace specialized architectures and task-tailored components of expert policies with large-scale data collection and setup-specific fine-tuning.
In this machine learning-focused workflow that is centered around models and scalable training, traditional robotics software frameworks become a bottleneck, while robot simulations offer only limited support for transitioning from and to real-world experiments.
In this work, we close this gap by introducing \emph{Robot Control Stack} (\rcs{}), a lean ecosystem designed from the ground up to support research in robot learning with large-scale generalist policies.
At its core, \rcs{} features a modular and easily extensible layered architecture with a unified interface for simulated and physical robots, facilitating sim-to-real transfer.
Despite its minimal footprint and dependencies, it offers a complete feature set, enabling both real-world experiments and large-scale training in simulation.
Our contribution is twofold: First, we introduce the architecture of \rcs{} and explain its design principles.
Second, we evaluate its usability and performance along the development cycle of VLA and RL policies.
Our experiments also provide an extensive evaluation of Octo, OpenVLA, and \pizero{} on multiple robots and shed light on how simulation data can improve real-world policy performance. Our code, datasets and videos are available at \href{https://robotcontrolstack.github.io/}{https://robotcontrolstack.github.io}
\end{abstract}

\bstctlcite{IEEEexample:BSTcontrol}

\nocite{shakey, octo, openvla, pizero, openx, droiddataset, robominddataset, yarp, ros, metaworld, robosuite, maniskill3, mujoco, gymnasium, roboticslibrary, roboticstoolbox, ompl, carpentier2019pinocchio, arkframework, pandapy, frankx, deoxyscontrol, polymetis, robosuite, isaaclab, lerobot}

\begin{table*}[t!]
\centering
\vspace{5.1pt}
\caption{Feature matrix of software packages for robot learning most closely related to \rcs{}}
\begin{tabularx}{\textwidth}{X|cc|cc|ccc|cc}
     \toprule
     & \multicolumn{2}{c|}{\textit{Robot Support}} & \multicolumn{2}{c|}{\textit{API}} & \multicolumn{3}{c|}{\textit{Scalability \& Deployment}} & \multicolumn{2}{c}{\textit{Robot Learning Features}} \\[0.1cm]
     \textbf{Name} & \textbf{Hardware} & \textbf{Simulation} & \textbf{Gymnasium} & \textbf{Low Level} & \textbf{Distributed} & \textbf{Parallel} & \textbf{pip} & \textbf{Online}   & \textbf{Offline (Data Collection)}  \\
     \midrule
     Isaac Lab \cite{isaaclab} & ROS2  & Isaac Sim & \cmark & \xmark & \cmark & \cmark & \cmark & \cmark & \cmark \\
     frankx \cite{frankx} & Panda & \xmark & \xmark & \xmark & \xmark & \xmark & \cmark & \xmark & \xmark \\
     LeRobot \cite{lerobot} & Universal & MuJoCo & \cmark & \xmark & \cmark & \cmark & \cmark & \cmark & \cmark \\
     Deoxys \cite{deoxyscontrol} & Panda & robosuite & \xmark & \xmark & \cmark & \xmark & \xmark & \xmark & \cmark \\
     Polymetis \cite{polymetis} & Panda & Multiple & \xmark & \xmark & \cmark & \xmark  & \cmark & \xmark & \xmark \\
     Ark \cite{arkframework} & Universal & Multiple & \cmark & C++ & \cmark & \xmark & \cmark & \xmark & \cmark  \\
     \midrule
     \textbf{RCS (ours)} & Universal & MuJoCo & \cmark & C++ & \cmark & \cmark & \cmark & \cmark & \cmark \\
     \bottomrule
\end{tabularx}
\label{tab:comparison}
\end{table*}

\section{Introduction}
\begin{figure}[htbp]\label{fig:eyecandy}
    \centering
    \includegraphics[width=0.7\linewidth]{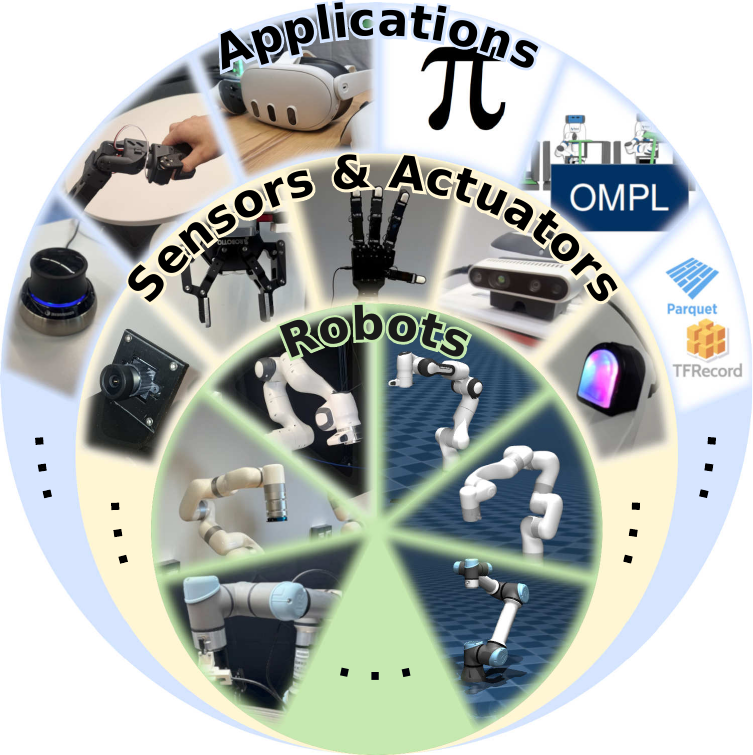}
        \caption{The layered architecture of \rcs{}: Applications at the top layer access robots, sensors, and actuators through a Gymnasium-based Python API for easy switching between hardware and a MuJoCo simulation. The lower layers expose a C++ API for performance-critical features, making RCS equally suitable for end-to-end policy learning and low-level control.
        }
    \label{fig:overview}
\end{figure}
Machine learning is playing an increasingly important role in robotics research.
While robotics research has early adopted methods from artificial intelligence~\cite{shakey}, recent developments in foundation models are challenging traditional approaches to robot learning and opening up new opportunities.
Vision-Language-Action models (\vla{}s) for robot manipulation combine multiple components, ranging from perception to action generation, into end-to-end policies capable of generalizing across tasks, robots, and environments~\cite{octo, openvla, pizero}.
Training them has only become possible due to community-wide data collection efforts~\cite{openx, droiddataset, robominddataset} that enable training at scale.

This approach differs fundamentally from traditional robotics experiments, where machine learning is often only a small part of the overall method and trained models become part of a predefined system architecture. 
Software packages for robotics are often designed based on this principle~\cite{yarp, ros} and thus do not fit well with machine learning-focused research, where the robot setup becomes part of the training and inference loop, not the other way around.
Many simulators targeted at robot learning solve this issue and scale well for highly parallelized training, but lack core robotics functionality and offer only limited support for controlling physical robots~\cite{metaworld,robosuite,maniskill3}.
As a result, research on \vla{}s and robot learning in general still requires customization for every new setup, model, and task.
What is still missing is a software ecosystem that flexibly adapts to the unique properties of individual robot setups, integrates well with community standards to enable sharing of models and data, and interfaces seamlessly with state-of-the-art machine learning tools and workflows to support research and development of new models and methods.

In this work, we propose \emph{Robot Control Stack (\rcs{})}, a library and tool set for robot learning that has been designed from the ground up for research on \vla{}s.
\rcs{} is based on a lightweight layered architecture that can be easily extended at different levels of abstraction.
This is illustrated in \autoref{fig:overview}.
At its core, \rcs{} provides C++ interfaces to support components with low-level APIs, and unify control between the integrated MuJoCo simulation~\cite{mujoco} and the actual robot hardware.
A complementary Gymnasium-based~\cite{gymnasium} Python API with modular environment wrappers enables the convenient implementation of high-level applications, such as data collection, deployment of trained policies, or traditional task and motion planning,
while the tight integration with MuJoCo enables seamless sim-to-real and real-to-sim support. 
In brief, \rcs{} is a lean ecosystem for robot learning, \emph{built to adapt to your application, rather than forcing you to adapt to it}.

\textbf{In summary, we make the following contributions:}
\begin{enumerate}
    \item  We introduce \rcs{} and show how its environment wrapper-based architecture enables adding new features at different levels of abstraction, supporting both Python- and C++-based robot software.
    \item We evaluate RCS on common use cases, including cross-embodiment support, training data collection in simulated and real environments, and the training and evaluation of \vla{} and RL agents.
    \item We provide extensive experimental results for Octo~\cite{octo}, OpenVLA~\cite{openvla}, and \pizero{}~\cite{pizero} on a reproducible picking task for a range of different robots.
    \item We highlight how mixing synthetic with real-world data can substantially improve the real-world performance of \pizero{}~\cite{pizero}.
\end{enumerate}

\section{Related Work}
Software packages for robotics have always played an important role in speeding up development and making research shareable.
Many libraries provide core robotics functionality~\cite{roboticslibrary, roboticstoolbox} and collections of algorithms for specific fields~\cite{ompl, carpentier2019pinocchio}.

Because typical robotics setups involve many components, assembling an integrated system quickly becomes complex.
This has motivated the development of robotics middleware like Yarp~\cite{yarp} and ROS \cite{ros}. While the former focuses on communication, ROS has grown into a large ecosystem that is widely used across research and industry. 
Although ROS is flexible enough to support practically any kind of architecture, building a robust, single-purpose stack requires understanding its complex underlying layers. Machine learning-driven applications must carefully coordinate multiple hardware and software components across both simulated and real environments. This is why task-optimized architectures typically incur less overhead and are easier to use than the more general and versatile ROS architecture.

Ark \cite{arkframework} implements an architecture similar to ROS, but adds essential machine learning features, such as a Gymnasium-based Python API~\cite{gymnasium}, native sim-to-real support, and data collection utilities.
However, it also lacks support for parallelization and for Reinforcement Learning (RL).
Unlike middleware-based systems, \rcs{} is not distributed by default, which enables synchronous execution if required, easy parallelization, and both online and offline learning.
Nevertheless, individual components can be distributed with Remote Procedure Calls (RPC).

In machine learning for robot manipulation, experimental setups are often simple, consisting only of a robot, a gripper, and cameras. Various libraries specifically designed for the popular Franka Emika Panda robot provide basic robot control functionality without the overhead of a large framework.
\mbox{panda-py}~\cite{pandapy} and frankx~\cite{frankx} offer a basic Python interface with a focus on motion generation, whereas the focus of Deoxys~\cite{deoxyscontrol} and Polymetis~\cite{polymetis} is on custom torque controllers.
The latter two also work with selected simulators.
In particular, Deoxys integrates with robosuite \cite{robosuite}.
\rcs{} also interfaces with both simulated and physical robots but is not limited to a single robot model.
In addition, it supports running a digital twin by running both the physical robot and the MuJoCo-based simulation~\cite{mujoco} in parallel.

Most related to our work are Isaac Lab~\cite{isaaclab} and \mbox{LeRobot}~\cite{lerobot}. Both are specifically designed for robot learning and tightly integrated into their respective ecosystems. Isaac Lab offers a comprehensive feature set for a wide range of applications and algorithms. However, while there is support for ROS, there is no support for seamlessly switching between simulation and hardware.
LeRobot, on the other hand, does not provide low-level API support, limiting hardware integration options. Moreover, it lacks core robotics functionalities such as Inverse Kinematics (IK) and path planning.

\autoref{tab:comparison} provides an overview of how \rcs{} compares with other software packages for robot learning. For clarity, only those with a scope similar to \rcs{} are included. In summary, \rcs{} offers a distinctive set of features that fills an important gap in robot learning.

\begin{figure*}[t]
  \centering
  \vspace{5.1pt}
  \includegraphics[width=\textwidth]{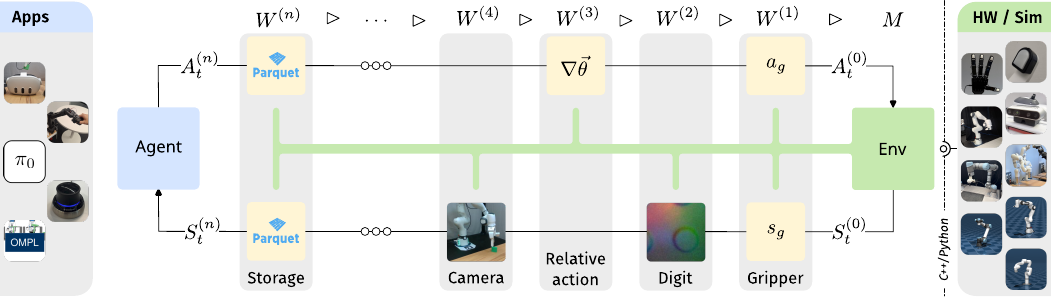}
  \caption{
    The architecture of \rcs{}. The applications on the left side interface with the environment on the right side, which can be a simulation or a real robot, through the Gymnasium interface. Sensors, actuators, and data observers wrap the environment by mutating the action, and/or the observation space.
  }
  \label{fig:arch}
\end{figure*}
\section{Methodology}\label{sec:meth}

\rcs{} is designed around the concept of environment wrappers.
An environment wrapper is a tuple $W = \langle f: S \to S^\prime, g: A^\prime \to A, P^\prime, R^\prime\rangle$, where $f$ and $g$ are mappings that transform the state and actions of a Markov Decision Process (MDP), $P^\prime$ and $R^\prime$ are, optionally, new state-transition probability function and reward function.
The wrapper $W$ can be \emph{applied} to an MDP $M = \langle S, A, P, R\rangle$ to produce a new MDP $M^\prime = \langle S^\prime, A^\prime, P^\prime, R^\prime \rangle$.
We denote $M^\prime = W \rhd M$ and say ``W wraps M''.
Note the difference in directionality between the state and action functions $f$ and $g$: $f$ transforms a state from the wrapped MDP $M$ to a state from the new MDP $M^\prime$, whereas $g$ transforms an action from the new MDP $M^\prime$ to the wrapped MDP $M$. They may also be implemented as identity functions $f_I$ and $g_I$, such that $S'_t = S_t = f_I(S_t)$ and $A_t^\prime = A_t = g_I(A_t)$.

The formalism can be extended to Partially Observable Markov Decision Processes (POMDPs) to better match the actual API of Gymnasium environments~\cite{gymnasium}.
The directionality of $f$ and $g$ remains unchanged, and we can reuse the relation $\rhd$ for denoting the wrapping of POMDPs.

\subsection{Library Architecture}

\autoref{fig:arch} shows the architecture of RCS.
At its lowest layer is a C++ interface that defines all functions needed to control a robot in an abstract manner. This interface has Python bindings, such that support for new robots can be implemented in both languages. The base environment is configured with a control type, e.g., Cartesian or joints, synchronous or asynchronous actions, and utilizes the interface functions. This design is implementation-agnostic and works with any robot or even a MuJoCo scene. 

Each scene is a sequence of $n$ wrappers, each of which can mutate the action and/or observation space of the environment. For example, a gripper wrapper adds an additional dimension to both action and observation spaces, a camera wrapper adds a camera frame to the observation space, etc.
Formally, at each time step, an agent, e.g.~a policy or a teleoperator, issues an action $A_t = A_t^{(n)}$ to the wrapped MDP $M^{(n)} =  W^{(n)} \rhd \dots \rhd W^{(1)} \rhd M$ that is then propagated through the action mutation function chain of the wrappers, i.e.~$A_t^{(0)} = g^{(1)}(\dots g^{(n)}(A_t^{(n)}))$. $A_t^{(0)}$ is then passed to the base MDP, i.e.~the interface that implements the robot that produces an observation state $S_t^{(0)}$. Analogous to the action, the observation mutation function chain updates the state: $S_t^{(n)} = f^{(n)}(\dots f^{(1)}(S_t^{(0)}))$ and is returned to the agent, and the process repeats until a termination state is reached. Identity wrappers can be used to record trajectory data or to stream the actions and observations over the network between a robot and a remote machine.

\subsubsection{Hardware Abstraction}
In principle, adding new hardware requires writing a new wrapper. However, since many hardware devices have a similar output and require similar utility code, e.g.~every camera outputs an RGB image and requires a polling thread, \rcs{} defines interfaces and off-the-shelf wrappers for common sensors and actuators.
These include a wrapper that implements polling for a set of cameras that follow a camera interface, and a wrapper for end effectors such as grippers or robot hands.

By default, Gymnasium environments are synchronous because the agent's action $A_t$ needs to be fully reflected by the environment's return state $S_{t+1}$. Therefore, RCS runs synchronously by default for both simulation and hardware. An action is only returned once the defined state has been reached. Still, it is possible to configure RCS to execute actions asynchronously such that the step function returns instantly, e.g. in teleoperation. Either way, the step-based design ensures that observations from different sensors are temporally synchronized.

\subsubsection{Simulation}
\rcs{} leverages the MuJoCo physics simulation and extends its API with customized functions for robotics use cases, leaving MuJoCo's core data structures exposed to the user for flexibility.
The aim is to provide an object-oriented view of a simulated scene for quick experimentation and to simplify adding new simulated robots, actuators, and sensors.

To enable synchronous operation and interrupts, \rcs{} implements a callback mechanism around MuJoCo's \textit{step} function.
For example, in a scene with a robot arm, a synchronous high-level method \textit{close gripper} requires the gripper to inform the simulation when it has reached its closed state.
The simulation should continue stepping until all entities have converged to a final state, e.g.~until the arm has reached its target position. 
In other cases, the simulation should be directly interrupted, e.g., when a collision is detected.
The callback mechanism enables individual simulation instances to implement such high-level functionality, mirroring traditional robotics software running on real hardware.

\subsubsection{Robotics Tool Kit}\label{sec:meth:toolkit}
\rcs{} also integrates established tools and libraries to provide core robotics functionality. 
We use Pinocchio~\cite{carpentier2019pinocchio} as a backend for kinematics.
With its broad compatibility with robot description file formats and low computational overhead, Pinocchio's functions have been integrated to work seamlessly across different robots in both simulated and real environments.
The robot model is based on the MuJoCo MJCF description, which can be read by Pinocchio.
Since \rcs{} ships with a MuJoCo base-scene of the supported robots, RCS offers a Pinocchio-based IK solver out of the box. OMPL~\cite{ompl} has also been integrated to allow for basic motion planning and more diversity in automated data collection tasks.

Since simulated and real-world robots use the same interface, it is easy to implement new tools that create synergies with each other, such as a parallel simulation instance that runs in real-time with the same actions that the real robot receives. In addition to its utility as a prototyping tool, it may be used in a planning context to verify an action before executing it on the real robot.
\rcs{} also includes a camera calibration interface that enables the implementation of custom setup-specific calibration strategies, such as using AprilTags, as shown in \autoref{fig:setups}.

\subsection{Applications and Ecosystem}
Applications (apps) are usually high-level scripts that utilize a use case- and setup-specific composition of the available wrappers to control a physical or simulated robot. Apps can also be wrappers that perform read-only operations, such as a storage wrapper that stores information from the scene in a specific format. Most apps fall into two categories: Generate and record robot behavior for downstream imitation learning or execute a policy for evaluation.
RCS is shipped with a wide variety of apps, such as teleoperation with different kinds of hardware, trajectory generators like OMPL, and policy inference engines.

VLA policies, and other recent deep learning-based models often come with a fairly rigid set of software dependencies, which can be incompatible with the dependencies of the software that interfaces with the target robotic platform. RCS provides a dedicated application layer library called \textit{VLAgents} to address this issue. VLAgents is implemented as a lightweight Python package with minimal dependencies that can be installed in the policy's Python environment. The VLAgents package is then used to interface with the policy's inference pipeline to create a server instance. This communicates with a client instance of VLAgents that is deployed in a parallel \rcs{} environment. Communication between the two instances is achieved through RPC, which uses TCP in the case of remote machines, or shared memory if both instances live on the same machine. This method has two main benefits: (1) RPC allows transmitted data to be accessed like a typical local variable, allowing a consistent interface on both sides; and (2) all observations in a given timestep come pre-aligned, as the alignment is handled by the layers that generate the data. 
VLAgents currently integrates seven different policies and three different simulation environments.

\section{Results}

In this section, we show that RCS can be used for the full pipeline of imitation and reinforcement learning with a series of experiments.
We describe how we collected and converted several datasets from both human and scripted sources, and how these datasets are used to conduct early-stage model evaluation in simulation and full-fledged real-world evaluation.

\begin{figure}[t]
  \centering
  \vspace{5.1pt}
  \includegraphics[width=.49\linewidth]{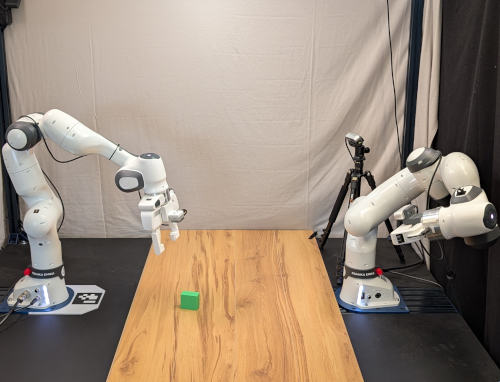}
  \hfill
  \includegraphics[width=.49\linewidth]{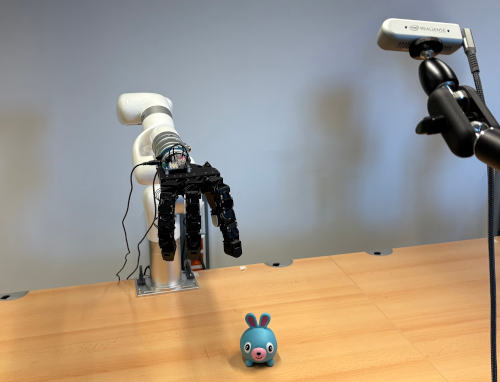}\\[6pt]
  \vfill
  \includegraphics[width=.49\linewidth]{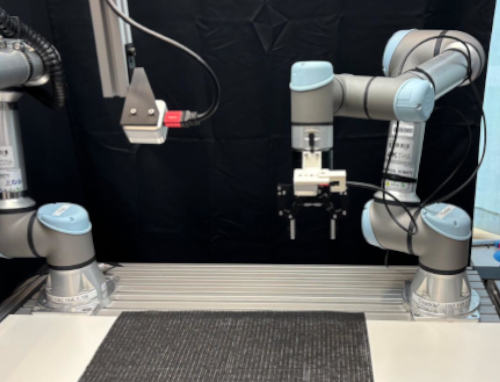}
  \hfill
  \includegraphics[width=.49\linewidth]{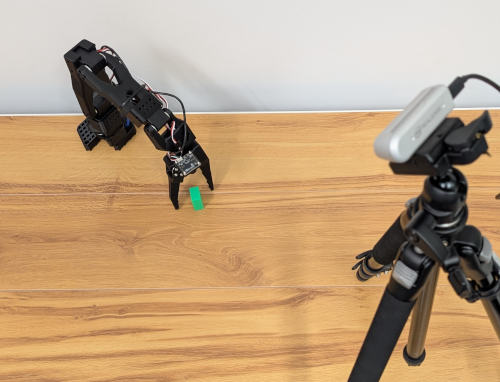}
  \caption{
  We deployed RCS in four different setups.
  \textit{From top left to bottom right:}
  \textbf{FR3} with Franka Hand gripper, wrist and side cameras;
  \textbf{xArm7} with Tilburg Hand and one side camera;
  \textbf{UR5e} with Robotiq gripper, wrist and side cameras;
  \textbf{SO101} with wrist and side cameras.
  }
  \label{fig:setups}
\end{figure}

\begin{table}[t]
\caption{Components implemented in \rcs{}. Bold font indicates availability in hardware and simulation.}
\label{tab:hardware}
\begin{tabularx}{\linewidth}{X|l|X|l}
\toprule
\textbf{Robots} & \textbf{End effectors} & \textbf{Sensors} & \textbf{User Input} \\
\midrule
- \textbf{\franka{}}      & - \textbf{Franka Hand}      & - \textbf{D400s}         & - HTC Vive       \\
- \textbf{\xarm{}}        & - \textbf{Tilburg Hand}        & - \textbf{Webcams}            & - Meta Quest 3   \\
- \textbf{\ur{}}          & - \textbf{Robotiq 2F85}     & - DIGIT~\cite{digit}      & - SpaceMouse     \\
- \textbf{\so{}}          & - \textbf{SO101 Gripper}  & - Tacto~\cite{tacto}  & - Keyboard   \\
\bottomrule
\end{tabularx}
\vspace{-1em}
\end{table}

\begin{figure}[t]
    \vspace{5.1pt}
	\centering
	\includegraphics[width=0.9\linewidth]{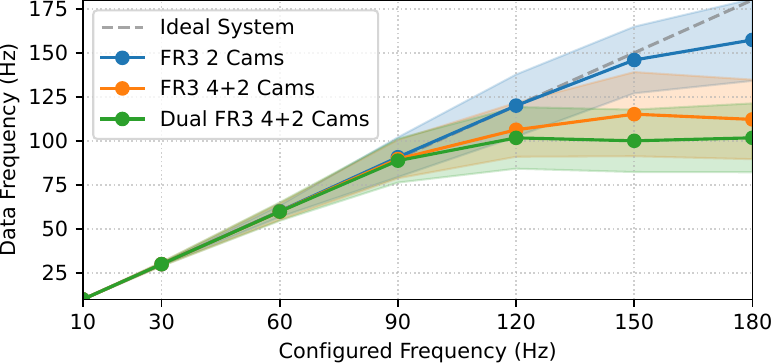}
	\caption{
    Configured control vs. measured data frequency during teleoperation, averaged over more than 1000 steps. The shaded area denotes the standard deviation. \emph{FR3 2 Cams}: Two RealSense cameras. \emph{FR3 4+2 Cams}: Four RealSense cameras and two DIGIT sensors. \emph{Dual FR3 4+2 Cams}: Like \emph{FR3 4+2 Cam} but with two FR3 robots.
    }
	\label{fig:latency}
    \vspace{-1em}
\end{figure}

\subsection{Deployed Setups} 
\autoref{tab:hardware} shows the list of components implemented in RCS. Four combinations of these components are set up across three different institutions. We refer to these setups using the name of their respective robotic arm: \franka{}, \xarm{}, \ur{} and \so{}.
They are shown in \autoref{fig:setups}.
In addition to the physical setups, we created a MuJoCo scene which replicates the setup of the FR3, shown in \autoref{fig:real_simified}.

\subsubsection{Hardware}
The \emph{\franka{}} setup consists of one Franka Research 3 (\franka{}), a 7-degree-of-freedom (DoF) robot arm equipped with a Franka Hand gripper, two cameras, and four fixed lights.
The first camera is a RealSense D405 mounted on the robot's wrist, and the second camera is a RealSense D435 mounted on a second \franka{} for reproducible camera poses, providing a third-person perspective of the scene.
Both RealSense cameras are calibrated with respect to the robot base and return RGB and depth images.
The \emph{\xarm{}} setup consists of a 7-DoF \xarm{} robotic arm, equipped with a Tilburg Hand, a 16-DoF multi-fingered robotic hand.
A single RealSense D435i side camera provides a view of the experiment environment from the front of the setup.
The \emph{\ur{}} setup consists of a 6-DoF \ur{} robotic arm, equipped with a Robotiq 2F-85 two-finger gripper. The setup includes two RealSense D405 cameras. One is a wrist camera and the other is a static camera positioned overhead, looking down at the scene.
The \emph{\so{}} consists of a 5-DoF \so{} robotic arm with its built-in 1-DoF gripper. One InnoMaker USB webcam is mounted to its wrist, and a single \mbox{RealSense D435} provides a diagonal view of the scene.

\subsubsection{Simulation}
\begin{figure}[t]
  \vspace{5.1pt}
  \centering
    \includegraphics[width=0.98\linewidth]{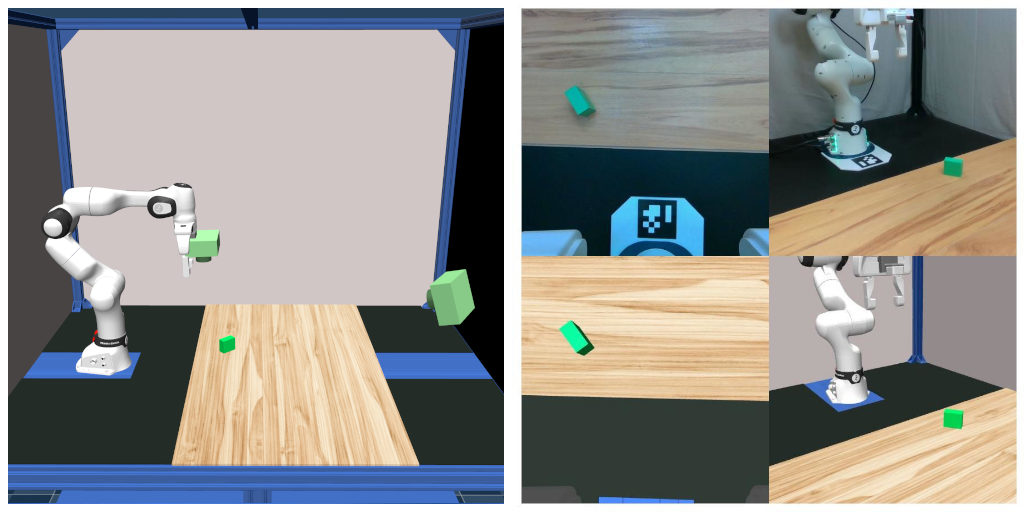}
        \caption{Replicated simulation scene. \textit{Left:} The simulation scene of \franka{}, with calibrated camera poses visible in light green. \textit{Right:} Images from the wrist and side cameras in real setup (above) vs. in simulation (below).}
        \label{fig:real_simified}
        \vspace{-1em}
\end{figure}

Every robot integrated in RCS ships with a default MuJoCo scene for easy experimentation. 
For our VLA sim-to-real experiments, we reproduced the \franka{} scene in simulation to visually match the setup as closely as possible.
We used the CAD file for our custom table, positioned the \franka{} at the same measured coordinates, and matched the background, table color, and wood texture to our lab setup.
Using the extrinsic calibration process described in~\autoref{sec:meth}, we place the two cameras at the same position in the simulated scene.
We also matched the intrinsic parameters and resolution of our RealSense cameras.
\autoref{fig:real_simified} shows both the real and the matched MuJoCo scene. 

\subsection{System Performance}
While early VLA architectures, such as OpenVLA, run at rather low inference frequencies of around \qty{5}{Hz}, more recent models, such as $\pi_0$, can achieve \qty{50}{Hz}. To validate whether \rcs{} meets these performance requirements, we measured the data recording frequency at fixed control frequencies during teleoperation across the three setups described in \autoref{fig:latency}. All setups are based on up to two FR3 robot arms and include up to four RealSense 405 and 435d cameras set to a resolution of $1280\times 720$. Two of the setups additionally record from two DIGIT tactile sensors with a resolution of $320\times 240$. All cameras run at \qty{30}{fps}. If the recording frequency is higher, we use the latest frame available. This is not required for proprioceptive data, which is available at high frequencies.

\autoref{fig:latency} shows that all configurations achieve up to \qty{90}{Hz}, with the two-camera \franka{} reaching \qty{120}{Hz}. At higher values, IO bottlenecks cause jitter and the maximum frame rate is limited by the sensor count. Importantly, \rcs{} scales well even at higher frequencies despite the synchronous nature of Gymnasium environments, making it suitable for modern VLA and RL deployments.

\subsection{Task and Dataset}
In imitation learning, the dataset---recordings of a robot performing a task---is crucial for downstream model performance.
In order to compare different visuomotor policies, i.e. VLA policies, we require datasets collected on a benchmark task. 
The criteria for the benchmark task are: (1) it should be executable in both real-world and simulation; and (2) it should be reproducible across setups.
While there are existing benchmarks that fulfill these criteria, such as FurnitureBench~\cite{heo2023furniturebench}, they often contain tasks that are too complex for less capable models, preventing meaningful comparisons between the models.
To make this possible, we designed a simple task that we refer to as \pickboxtask{}: grasping a green 3D-printed cuboid in various orientations. This cuboid is placed randomly within a defined area of the workspace. The random position and orientation are seeded to be reproducible. In simulation, the cuboid pose can be directly set. In a real-world setting, we place it at random manually or using the robot itself. See \autoref{fig:real_simified} for an example of a simulation and a real-world task scene.

For the recording, we use a recorder wrapper that takes the data from the scene, temporally aligns observations and actions, and writes the data into a Parquet file. All of our data was recorded at a frequency of \qty{30}{Hz}.
The observations in our datasets contain the full robot state, including absolute joint angles in radians, Cartesian end effector pose in the robot base frame (coordinate system is unified among all robots), gripper state, and camera RGB and depth images. 
Optionally, if the camera is calibrated, the camera data frames also contain intrinsic and extrinsic calibration matrices.

In the following paragraphs, we give a brief description of how the dataset was collected with each setup.

\paragraph{\franka{}}
We teleoperated 143 demonstrations of the \pickboxtask{} task on the \franka{} setup using the HTC Vive VR input device with an operational space controller.
As outlined before, we used the robot arm itself to place the cuboid to a random position and orientation within a $\qty{30}{cm}\times\qty{40}{cm}$ workspace.
With this strategy, we can sample the position of the cube within a defined region unbiased from human placement. In addition, we also know the position of the cube, which is recorded in the dataset and can be used to replay the trajectory in simulation.

\paragraph{\xarm{}}
We collected 100 demonstrations from teleoperation with a Meta Quest VR controller.
Since the cuboid in the \pickboxtask{} task was too small for reliable grasping with the Tilburg Hand, we substituted it with a deformable toy that better fits the size of the hand.
We refer to this modified task as the \picktoytask{}.
For each trial, we manually place the object in a random position and orientation within a $\qty{60}{cm}\times\qty{30}{cm}$ workspace.
We then teleoperated the robot to reach the object from the home position, grasp it, and lift it to the home position.

\paragraph{\ur{}}
We used a SpaceMouse for Cartesian velocity control to teleoperate the robot and collect 167 demonstrations. For each demonstration, a cuboid of the \pickboxtask{} was manually placed in a random position and orientation within a $\qty{50}{cm}\times\qty{50}{cm}$ workspace.
We teleoperated the robot to grasp it,
lift it, and then return to the home position while holding the object.

\paragraph{\so{}}
We followed the same experiment procedure as in the other setups, but used a smaller cuboid in the \pickboxtask{} to match the robot's size, and a workspace of $\qty{35}{cm}\times\qty{20}{cm}$. In total, we collected 120 demonstrations using the robot's leader-follower teleoperation setup.

\paragraph{Scripted Simulated \franka{}}
A scripted simulation episode is generated by using a simple linear interpolator to guide the robot's end effector to the cuboid, whose pose is contained in the MuJoCo data. A simple wrapper that observes the height of the cuboid was used to determine the success of the task, since success is not guaranteed due to contact simulation uncertainties. This process can be multi-threaded to yield around 1000 episodes per 20 minutes on a consumer-grade GPU-equipped laptop, and then filtered based on the success criterion. Using the simulation environment described above and the same task configuration as the real \franka{}, we generated 3000 such episodes with a success rate of 73\%, resulting in a dataset containing 2193 successful simulation demonstrations.

To qualitatively explore the gap between the real-world and the simulated \franka{} setup, we used the observed robot states from the \franka{} demonstrations described above and replayed every episode in simulation. Thanks to the automated cuboid placement algorithm and camera calibration, we were able to replicate the scene closely, as shown in \autoref{fig:real_simified}. The replayed dataset was filtered using the success criterion described above, resulting in 73 successful episodes.

\subsection{VLA Experiments}
 To show how \rcs{} supports and accelerates VLA research, we evaluate different VLAs on the robot setups on the presented tasks.

\subsubsection{\pizero{} on Different Setups}

\begin{figure}[t]
  \centering
  \vspace{5.1pt}
  \includegraphics[width=\linewidth]{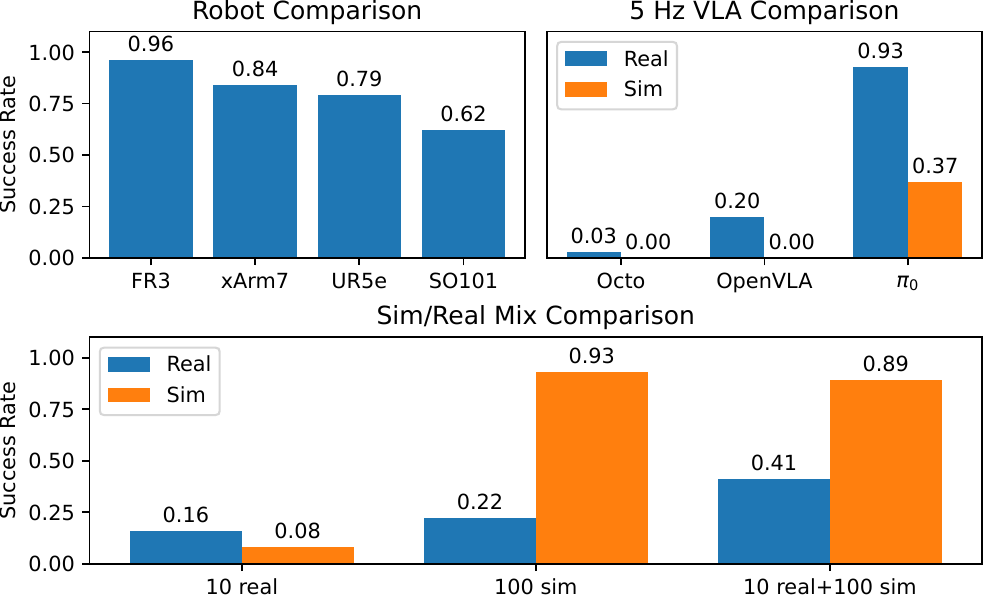}
  \caption{
  VLA success rates in different settings.
  \textit{Top left:} 
  \pizero{} fine-tuned on four datasets from different setups.
  \textit{Top right:} 
  Octo, OpenVLA and \pizero{} fine-tuned
  on our \franka{} setup (real) with a down-sampled frequency of \qty{5}{Hz} and evaluated on the real-world setup and the replicated simulated scene.
  \textit{Bottom:}
  Different data mixes of synthetic and real data evaluated on the real-world setup and the simulated scene.
  The number in the experiment name denotes how many episodes from the respective domain were used in the training mix.
  }
  \label{fig:bar_plots}
  \vspace{-1em}
\end{figure}

We evaluate \pizero{} across all four setups introduced earlier.
We fine-tune a model for each setup on the datasets described above.
Each model uses its own weights, and each experiment is independent.
Nevertheless, due to the flexible wrapper-based architecture of \rcs{}, we can share the code for teleoperation and model execution, minimizing hardware-induced overhead.
For the evaluation, we used the same \qty{30}{Hz} control frequency as we did during data collection, and employed action chunking with a prediction horizon of 20 steps. Each model is evaluated on 50 real-world rollouts.
The resulting evaluation success rates are shown in the top left plot of \autoref{fig:bar_plots}.

\paragraph{\franka{}}
Out of all hardware platforms, the \franka{} setup achieves the highest performance with \pizero{}.
This result is expected since some portion of \pizero{}'s pre-training dataset contains the Franka robotic arm.

\paragraph{\xarm{}}
On the \xarm{} setup, \pizero{} performs well despite the substantial difference in the end effectors used compared to other setups, highlighting its ability to generalize across embodiments, including multi-fingered robotic hands.
This performance may also be partly attributable to the inherent robustness of multi-fingered grasping, and the grasp-friendly geometric and physical features of the object used in the \xarm{} setup for the \picktoytask{} task.
Since the object does not require to be grasped from a specified angle, unlike the green cuboid, the task is also simpler.

\paragraph{\ur{}}
The \ur{} robot and Robotiq gripper were also part of \pizero{}'s pre-training dataset, hence the good result. One reason for the performance gap to the \franka{} may be the larger workspace used in the \ur{} experiments, which complicates the task. This applies in particular in the vicinity of robot singularities. Additionally, in the \ur{} experiments, the cuboid was placed manually, implying some human bias.

\paragraph{\so{}}
\pizero{} performs noticeably poorer on the \so{} compared to other robots, which we assume to be due to following factors: (1) The robot's lower DoF and much smaller size do not align well with the datasets that \pizero{} was trained on. (2) Picking up objects on the gripper's fixed finger side frequently results in unintended collisions, while objects on the movable finger's side suffer less from this issue. (3) The robot's inherent low-cost components lead to higher margins of error during control, such as link deflections and motor backlash.

\subsubsection{Different VLAs on Next Step Prediction}
In this experiment, we compare three different open-source VLA policies on the \franka{} setup, namely versions of \octo{}, \ovla{}, and \pizero{} that were fine-tuned on the \franka{} dataset.
These models represent recent developments in open-source VLAs.
Since the models differ in their architecture with respect to action prediction, we downsample our data to \qty{5}{Hz}, the frequency used in \ovla{}, and only perform synchronous next step prediction.
This is because \ovla{} can only do next-step prediction and, thus, it would be unfair to compare it against models that perform action chunking. Moreover, higher frequencies can be problematic for next-step prediction as the model sees almost the same state as before and gets stuck in small movements. Finally, the models have been trained to predict exact states. Thus, synchronous operation should yield the best results in this case.

We evaluate the models for the same cuboid placement positions over 30 rollouts on the real-world setup and over 100 rollouts on the replicated MuJoCo scene (see \autoref{fig:real_simified}). The latter evaluation sheds light on the real-to-sim capabilities, similar to SIMPLER~\cite{simpler}.
Strong real-to-sim performance not only shows the generalization of the model, but is also an important metric if it correlates with the real-world performance, as it can be used during training for model design and hyperparameter selection.
Evaluation results are shown in the top right plot of~\autoref{fig:bar_plots}.

\octo{} and \ovla{} show low performance even for this rather easy picking task.
This can be explained by the distribution of the pre-training dataset, which has only a small fraction of data with the Franka Emika Panda robot, the predecessor of the \franka{}.
Low success rates of these models are also in line with previous work~\cite{pizero}.
The performance of \pizero{} decreases slightly compared to its \qty{30}{Hz} version, which is expected since the pre-training dataset mainly contains \qty{50}{Hz} and \qty{30}{Hz} trajectories.
It is also capable of some degree of domain transfer to simulation, whereas the other two models do not generalize to the new simulation domain.

\subsubsection{\pizero{} Evaluation in Simulation}
Similar to the idea in SIMPLER, we evaluate the \pizero{} policy fine-tuned on our \franka{} dataset in simulation (real-to-sim).
Instead of evaluating only the end result, as done in the previous section, we evaluate each checkpoint throughout the training.
The resulting success rate plot is shown in the left part of~\autoref{fig:success_rate_sim_real}.
As shown in the figure, the success rate in the simulated environment provides a lower bound for the real-world setup, which may be due to the domain shift induced by the simulation.
In line with SIMPLER, our results indicate a loose correlation between simulation and the real domain.

\subsubsection{\pizero{} with Synthetic Data Mixes}
In this experiment, we train \pizero{} on different data mixes of real and synthetic data from the \franka{} and the scripted simulated \franka{} datasets.
The bar plot at the bottom in~\autoref{fig:bar_plots} shows the three training mixes: One with 10 episodes from the real domain, one with 100 episodes from the simulated domain and one with a combination of both.
The first experiment shows that \pizero{} does not transfer to a new task and setup with a small dataset. Interestingly, it is still capable of real-to-sim transfer.
The results of the second experiment suggest that this capability is also present in the sim-to-real direction.
As expected, the in-distribution evaluation in simulation exceeds the real-world performance.
Interestingly, when fine-tuning on a mix of both simulated and real-world data, the performance increases over-proportionally, even though only 10 episodes of real-world demonstrations are provided in the training set.

\begin{figure}[t]
  \centering
  \vspace{5.1pt}
  \includegraphics[width=\linewidth]{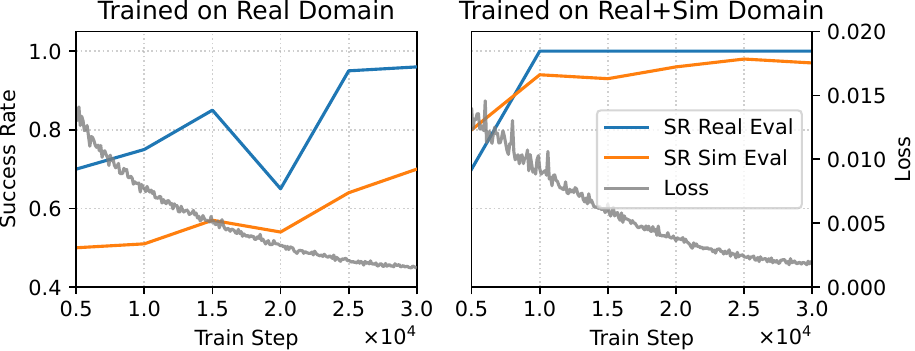}
  \caption{
  Evaluation success rates of all training checkpoints. Every checkpoint is evaluated on 20 real and 100 simulated rollouts. Left: Trained on 143 episodes on the \franka{} dataset. Right: Trained on a mix of 143 episodes from the \franka{} dataset and 500 episodes from the scripted dataset of the replicated simulated domain.
  }
  \label{fig:success_rate_sim_real}
  \vspace{-1em}
\end{figure}

We also observe this effect with a data mix of 143 real-world and 500 simulated episodes. As shown in the right part of \autoref{fig:success_rate_sim_real}, the model achieves a perfect success rate already after 10,000 steps of training.
In addition, the simulation performance is closer to the real-world performance compared to the training with real-world data only, as the evaluation domain is in-distribution.
The success rate in simulation is lower, which can be explained by the fact that the task is harder: The cuboid must be grasped close to the center in order for the contact simulation to work properly. By contrast, in the real-world experiment, grasps that are close to the edges often still work.

\subsection{RL Training Pipeline}
The requirements for RL and VLA training pipelines differ considerably.
VLA training is usually GPU-bound, involving a standard deep learning pipeline where a large neural network is trained on a dataset.
In contrast, RL training involves tight interaction loops between simulation and policy inference.
While recent GPU-based simulators enable large-scale parallelism with considerably higher throughput, vision-based policies and the associated rendering costs account for a substantial fraction of the overall computation costs. As a result, a CPU-based MuJoCo simulation offers a good trade-off between performance and broad usability.
With this experiment, we show how \rcs{} caters to the needs of RL pipelines by training a policy to solve the \pickboxtask{} in the replicated MuJoCo scene shown in \autoref{fig:real_simified}.

RCS natively supports parallelization, and parallel training works out of the box with Stable Baselines 3 (SB3)~\cite{sb3}. Its Gymnasium-based API is compatible with the majority of Python-based RL libraries.
When rendering a single camera image at low resolution, our training pipeline runs at more than 2000 steps per second when scaling to 24 parallel environments. Note that a large portion of the runtime cost is due to rendering and can be significantly reduced by simplifying the scene.
This shows that \rcs{} is not a bottleneck in the RL training pipeline: the performance depends on the MuJoCo model and the RL algorithm's implementation. 

We use the PickCube-v1 reward function of \mbox{ManiSkill 3} for potential-based reward shaping~\cite{pbrs} and add a discrete success reward. The function is based on information---distances between objects and whether the cube is grasped---available via \rcs{}.
Using this reward, we train a vanilla PPO policy with SB3 using two RGB cameras and proprioceptive state as inputs, and relative joint angles as output.
Neither performance nor hyperparameters were tuned. The policy learns to lift the cube with a 100\% success rate within three hours (8.5M environment steps) on an Nvidia RTX~4080 and a 12-core CPU. This demonstrates that the robotics-focused tool set of \rcs{} supports the development of RL environments and the design of reward functions.

\section{Conclusion and Future Work}
In this work, we introduced \rcs{}, a lean ecosystem designed for scalable robot learning that integrates seamlessly with machine learning workflows.
Different from other robot learning frameworks, \rcs{} offers a full feature set at minimal overhead but maximum flexibility and performance.
Our experiments show that \rcs{} makes it easy to implement data collection, evaluation, and inference for robot learning on different types of robots and in simulation. This enabled us to collaboratively perform an extensive evaluation of selected \vla{}s on a picking task and show how the inclusion of simulation data can boost policy performance.

The initial release of \rcs{} is only a starting point for the ecosystem. We are planning to provide an interface to ROS and to add support for bimanual and mobile manipulation tasks.
Combined with the already existing support for tactile sensing, this will make \rcs{} a future-proof ecosystem for research in humanoid robotics.

\bibliographystyle{IEEEtran}
\bibliography{bibliography.bib}

@IEEEtranBSTCTL{IEEEexample:BSTcontrol,
CTLuse_forced_etal = "yes",
CTLmax_names_forced_etal = "6",
CTLnames_show_etal = "3",
CTLdash_repeated_names = "no"
}

@string{SII = "IEEE Int.~Symp.~on System Integrations (SII)"}

@string{ieeeral = "IEEE Robotics and Automation Letters"}

@string{IROS = "Proc.~of the IEEE/RSJ Int.~Conf.~on Intelligent Robots and Systems (IROS)"}

@string{ICRA = "Proc.~of the IEEE Int.~Conf.~on Robotics \& Automation (ICRA)"}

@string{IJRR = "International Journal of Robotics Research"}

@string{ICML = "Proc.~of the Int.~Conf.~on Machine Learning (ICML)"}

@string{IJRR = "Int.~Journal of Robotics Research (IJRR)"}

@string{rss = "Proc.~of Robotics: Science and Systems (RSS)"}

@string{jmlr = "Journal of Machine Learning Research"}

@string{CORL = "Proc.~of the Conf.~on Robot Learning (CoRL)"}

@inproceedings{roboticstoolbox,
  author={Corke, Peter and Haviland, Jesse},
  booktitle=ICRA, 
  title={Not your grandmother’s toolbox – the Robotics Toolbox reinvented for Python}, 
  year={2021}
}

@inproceedings{roboticslibrary,
  author={Rickert, Markus and Gaschler, Andre},
  booktitle=IROS, 
  title={Robotics library: An object-oriented approach to robot applications}, 
  year={2017}
}

@inproceedings{deoxyscontrol,
  title = 	 {VIOLA: Imitation Learning for Vision-Based Manipulation with Object Proposal Priors},
  author =       {Zhu, Yifeng and Joshi, Abhishek and Stone, Peter and Zhu, Yuke},
  booktitle = 	 CORL,
  year = 	 {2023}
}

@misc{polymetis,
  author =       {Lin, Yixin and Wang, Austin S. and Sutanto, Giovanni and Rai, Akshara and Meier, Franziska},
  title =        {Polymetis},
  howpublished = {\url{https://facebookresearch.github.io/fairo/polymetis/}},
  year =         {2021}
}

@misc{frankx,
    author = {Berscheid, Lars},
    license = {LGPL-3.0},
    title = {{frankx: High-Level Motion Library for the Franka Emika Robot}},
    howpublished = {\url{https://github.com/pantor/frankx}}
}

@article{pandapy,
    title = {Taming the Panda with Python: A powerful duo for seamless robotics programming and integration},
    journal = {SoftwareX},
    year = {2023},
    author = {Jean Elsner}
}

@misc{ros,
  title={ROS: an open-source Robot Operating System},
  author = {Quigley, Morgan and Gerkey, Brian and Conley, Ken and Faust, Josh and Foote, Tully and Leibs, Jeremy and Berger, Eric and Wheeler, Rob and Ng, Andrew},
  booktitle={ICRA 2009 Open Source Software in Robotics},
  year={2009},
}

@article{yarp,
    author = {Giorgio Metta and Paul Fitzpatrick and Lorenzo Natale},
    title ={YARP: Yet Another Robot Platform},
    journal = {International Journal of Advanced Robotic Systems},
    year = {2006}
}

@misc{arkframework,
    title={Ark: An Open-source Python-based Framework for Robot Learning}, 
    author={Magnus Dierking and Christopher E. Mower and Sarthak Das and Huang Helong and Jiacheng Qiu and Cody Reading and Wei Chen and Huidong Liang and Huang Guowei and Jan Peters and Quan Xingyue and Jun Wang and Haitham Bou-Ammar},
    year={2025},
    howpublished={\url{https://arxiv.org/abs/2506.21628}}, 
}

@misc{lerobot,
    author = {Cadene, Remi and Alibert, Simon and Soare, Alexander and Gallouedec, Quentin and Zouitine, Adil and Palma, Steven and Kooijmans, Pepijn and Aractingi, Michel and Shukor, Mustafa and Aubakirova, Dana and Russi, Martino and Capuano, Francesco and Pascal, Caroline and Choghari, Jade and Moss, Jess and Wolf, Thomas},
    title = {{LeRobot}: State-of-the-art Machine Learning for Real-World Robotics in Pytorch},
    howpublished = {\url{https://github.com/huggingface/lerobot}},
    year = {2024}
}

@article{isaaclab,
    author={Mittal, Mayank and Yu, Calvin and Yu, Qinxi and Liu, Jingzhou and Rudin, Nikita and Hoeller, David and Yuan, Jia Lin and Singh, Ritvik and Guo, Yunrong and Mazhar, Hammad and Mandlekar, Ajay and Babich, Buck and State, Gavriel and Hutter, Marco and Garg, Animesh},
    journal=ieeeral, 
    title={Orbit: A Unified Simulation Framework for Interactive Robot Learning Environments}, 
    year={2023},
    volume={8},
    number={6}
}

@inproceedings{mujoco,
  author={Todorov, Emanuel and Erez, Tom and Tassa, Yuval},
  booktitle= IROS, 
  title={MuJoCo: A physics engine for model-based control}, 
  year={2012}
}

@misc{robosuite,
      title={robosuite: A Modular Simulation Framework and Benchmark for Robot Learning}, 
      author={Yuke Zhu and Josiah Wong and Ajay Mandlekar and Roberto Martín-Martín and Abhishek Joshi and Kevin Lin and Abhiram Maddukuri and Soroush Nasiriany and Yifeng Zhu},
      year={2025},
      howpublished={\url{https://arxiv.org/abs/2009.12293}}, 
}

@inproceedings{metaworld,
  title = 	 {{Meta-World}: A Benchmark and Evaluation for Multi-Task and Meta Reinforcement Learning},
  author =       {Yu, Tianhe and Quillen, Deirdre and He, Zhanpeng and Julian, Ryan and Hausman, Karol and Finn, Chelsea and Levine, Sergey},
  booktitle = CORL,
  year = 	 {2020}
}

@inproceedings{maniskill3,
    author = {Stone, Tao and Xiang, Fanbo and Shukla, Arth and Qin, Yuzhe and Hinrichsen, Xander and Yuan, Xiaodi and Bao, Chen and Lin, Xinsong and Liu, Yulin and Chan, Tse-Kai and Gao, Yuan and Li, Xuanlin and Mu, Tongzhou and Xiao, Nan and Gurha, Arnav and N., Viswesh and Choi, Yong Woo and Chen, Yen-Ru and Huang, Zhiao and Calandra, Roberto and Chen, Rui and Luo, Shan and Su, Hao},
    title = {Demonstrating GPU Parallelized Robot Simulation and Rendering for Generalizable Embodied AI with {ManiSkill3}},
    booktitle = rss,
    year = 2025
}

@article{tacto,
  author   = {Wang, Shaoxiong and Lambeta, Mike and Chou, Po-Wei and Calandra, Roberto},
  title    = {{TACTO}: A Fast, Flexible, and Open-source Simulator for High-resolution Vision-based Tactile Sensors},
  journal  = ieeeral,
  year     = {2022},
  volume   = {7},
  number   = {2}
}

@article{digit,
  author={Lambeta, Mike and Chou, Po-Wei and Tian, Stephen and Yang, Brian and Maloon, Benjamin and Most, Victoria Rose and Stroud, Dave and Santos, Raymond and Byagowi, Ahmad and Kammerer, Gregg and Jayaraman, Dinesh and Calandra, Roberto},
  title={{DIGIT}: A Novel Design for a Low-Cost Compact High-Resolution Tactile Sensor With Application to In-Hand Manipulation}, 
  journal=ieeeral, 
  year={2020},
  volume={5},
  number={3}
}

@misc{gymnasium,
      title={Gymnasium: A Standard Interface for Reinforcement Learning Environments}, 
      author={Mark Towers and Ariel Kwiatkowski and Jordan Terry and John U. Balis and Gianluca De Cola and Tristan Deleu and Manuel Goulão and Andreas Kallinteris and Markus Krimmel and Arjun KG and Rodrigo Perez-Vicente and Andrea Pierré and Sander Schulhoff and Jun Jet Tai and Hannah Tan and Omar G. Younis},
      year={2024},
      howpublished={\url{https://arxiv.org/abs/2407.17032}}
}

@inproceedings{openx,
  author={O’Neill, Abby and Rehman, Abdul and Maddukuri, Abhiram and Gupta, Abhishek and Padalkar, Abhishek and Lee, Abraham and Pooley, Acorn and Gupta, Agrim and Mandlekar, Ajay and Jain, Ajinkya and Tung, Albert and Bewley, Alex and Herzog, Alex and Irpan, Alex and Khazatsky, Alexander and Rai, Anant and Gupta, Anchit and Wang, Andrew and Singh, Anikait and Garg, Animesh and Kembhavi, Aniruddha and Xie, Annie and Brohan, Anthony and Raffin, Antonin and Sharma, Archit and Yavary, Arefeh and Jain, Arhan and Balakrishna, Ashwin and Wahid, Ayzaan and Burgess-Limerick, Ben and Kim, Beomjoon and Schölkopf, Bernhard and Wulfe, Blake and Ichter, Brian and Lu, Cewu and Xu, Charles and Le, Charlotte and Finn, Chelsea and Wang, Chen and Xu, Chenfeng and Chi, Cheng and Huang, Chenguang and Chan, Christine and Agia, Christopher and Pan, Chuer and Fu, Chuyuan and Devin, Coline and Xu, Danfei and Morton, Daniel and Driess, Danny and Chen, Daphne and Pathak, Deepak and Shah, Dhruv and Büchler, Dieter and Jayaraman, Dinesh and Kalashnikov, Dmitry and Sadigh, Dorsa and Johns, Edward and Foster, Ethan and Liu, Fangchen and Ceola, Federico and Xia, Fei and Zhao, Feiyu and Stulp, Freek and Zhou, Gaoyue and Sukhatme, Gaurav S. and Salhotra, Gautam and Yan, Ge and Feng, Gilbert and Schiavi, Giulio and Berseth, Glen and Kahn, Gregory and Wang, Guanzhi and Su, Hao and Fang, Hao-Shu and Shi, Haochen and Bao, Henghui and Ben Amor, Heni and Christensen, Henrik I and Furuta, Hiroki and Walke, Homer and Fang, Hongjie and Ha, Huy and Mordatch, Igor and Radosavovic, Ilija and Leal, Isabel and Liang, Jacky and Abou-Chakra, Jad and Kim, Jaehyung and Drake, Jaimyn and Peters, Jan and Schneider, Jan and Hsu, Jasmine and Bohg, Jeannette and Bingham, Jeffrey and Wu, Jeffrey and Gao, Jensen and Hu, Jiaheng and Wu, Jiajun and Wu, Jialin and Sun, Jiankai and Luo, Jianlan and Gu, Jiayuan and Tan, Jie and Oh, Jihoon and Wu, Jimmy and Lu, Jingpei and Yang, Jingyun and Malik, Jitendra and Silvério, João and Hejna, Joey and Booher, Jonathan and Tompson, Jonathan and Yang, Jonathan and Salvador, Jordi and Lim, Joseph J. and Han, Junhyek and Wang, Kaiyuan and Rao, Kanishka and Pertsch, Karl and Hausman, Karol and Go, Keegan and Gopalakrishnan, Keerthana and Goldberg, Ken and Byrne, Kendra and Oslund, Kenneth and Kawaharazuka, Kento and Black, Kevin and Lin, Kevin and Zhang, Kevin and Ehsani, Kiana and Lekkala, Kiran and Ellis, Kirsty and Rana, Krishan and Srinivasan, Krishnan and Fang, Kuan and Singh, Kunal Pratap and Zeng, Kuo-Hao and Hatch, Kyle and Hsu, Kyle and Itti, Laurent and Chen, Lawrence Yunliang and Pinto, Lerrel and Fei-Fei, Li and Tan, Liam and Fan, Linxi Jim and Ott, Lionel and Lee, Lisa and Weihs, Luca and Chen, Magnum and Lepert, Marion and Memmel, Marius and Tomizuka, Masayoshi and Itkina, Masha and Castro, Mateo Guaman and Spero, Max and Du, Maximilian and Ahn, Michael and Yip, Michael C. and Zhang, Mingtong and Ding, Mingyu and Heo, Minho and Srirama, Mohan Kumar and Sharma, Mohit and Kim, Moo Jin and Kanazawa, Naoaki and Hansen, Nicklas and Heess, Nicolas and Joshi, Nikhil J and Suenderhauf, Niko and Liu, Ning and Di Palo, Norman and Shafiullah, Nur Muhammad Mahi and Mees, Oier and Kroemer, Oliver and Bastani, Osbert and Sanketi, Pannag R and Miller, Patrick Tree and Yin, Patrick and Wohlhart, Paul and Xu, Peng and Fagan, Peter David and Mitrano, Peter and Sermanet, Pierre and Abbeel, Pieter and Sundaresan, Priya and Chen, Qiuyu and Vuong, Quan and Rafailov, Rafael and Tian, Ran and Doshi, Ria and Martín-Martín, Roberto and Baijal, Rohan and Scalise, Rosario and Hendrix, Rose and Lin, Roy and Qian, Runjia and Zhang, Ruohan and Mendonca, Russell and Shah, Rutav and Hoque, Ryan and Julian, Ryan and Bustamante, Samuel and Kirmani, Sean and Levine, Sergey and Lin, Shan and Moore, Sherry and Bahl, Shikhar and Dass, Shivin and Sonawani, Shubham and Song, Shuran and Xu, Sichun and Haldar, Siddhant and Karamcheti, Siddharth and Adebola, Simeon and Guist, Simon and Nasiriany, Soroush and Schaal, Stefan and Welker, Stefan and Tian, Stephen and Ramamoorthy, Subramanian and Dasari, Sudeep and Belkhale, Suneel and Park, Sungjae and Nair, Suraj and Mirchandani, Suvir and Osa, Takayuki and Gupta, Tanmay and Harada, Tatsuya and Matsushima, Tatsuya and Xiao, Ted and Kollar, Thomas and Yu, Tianhe and Ding, Tianli and Davchev, Todor and Zhao, Tony Z. and Armstrong, Travis and Darrell, Trevor and Chung, Trinity and Jain, Vidhi and Vanhoucke, Vincent and Zhan, Wei and Zhou, Wenxuan and Burgard, Wolfram and Chen, Xi and Wang, Xiaolong and Zhu, Xinghao and Geng, Xinyang and Liu, Xiyuan and Liangwei, Xu and Li, Xuanlin and Lu, Yao and Ma, Yecheng Jason and Kim, Yejin and Chebotar, Yevgen and Zhou, Yifan and Zhu, Yifeng and Wu, Yilin and Xu, Ying and Wang, Yixuan and Bisk, Yonatan and Cho, Yoonyoung and Lee, Youngwoon and Cui, Yuchen and Cao, Yue and Wu, Yueh-Hua and Tang, Yujin and Zhu, Yuke and Zhang, Yunchu and Jiang, Yunfan and Li, Yunshuang and Li, Yunzhu and Iwasawa, Yusuke and Matsuo, Yutaka and Ma, Zehan and Xu, Zhuo and Cui, Zichen Jeff and Zhang, Zichen and Lin, Zipeng},
  booktitle=ICRA, 
  title={{Open X-Embodiment}: Robotic Learning Datasets and {RT-X} Models}, 
  year={2024}
}

@inproceedings{octo,
    title={Octo: An Open-Source Generalist Robot Policy},
    author = {Ghosh, Dibya and Walke, Homer Rich and Pertsch, Karl and Black, Kevin and Mees, Oier and Dasari, Sudeep and Hejna, Joey and Kreiman, Tobias and Xu, Charles and Luo, Jianlan and Tan, You Liang and Chen, Lawrence Yunliang and Vuong, Quan and Xiao, Ted and Sanketi, Pannag R and Sadigh, Dorsa and Finn, Chelsea and Levine, Sergey},
    booktitle = RSS,
    year = {2024}
}

@InProceedings{openvla,
  title = 	 {OpenVLA: An Open-Source Vision-Language-Action Model},
  author =       {Kim, Moo Jin and Pertsch, Karl and Karamcheti, Siddharth and Xiao, Ted and Balakrishna, Ashwin and Nair, Suraj and Rafailov, Rafael and Foster, Ethan P and Sanketi, Pannag R and Vuong, Quan and Kollar, Thomas and Burchfiel, Benjamin and Tedrake, Russ and Sadigh, Dorsa and Levine, Sergey and Liang, Percy and Finn, Chelsea},
  booktitle = 	 CORL,
  year = 	 {2025},
}

@misc{pizero,
      title={$\pi_0$: A Vision-Language-Action Flow Model for General Robot Control}, 
      author={Kevin Black and Noah Brown and Danny Driess and Adnan Esmail and Michael Equi and Chelsea Finn and Niccolo Fusai and Lachy Groom and Karol Hausman and Brian Ichter and Szymon Jakubczak and Tim Jones and Liyiming Ke and Sergey Levine and Adrian Li-Bell and Mohith Mothukuri and Suraj Nair and Karl Pertsch and Lucy Xiaoyang Shi and James Tanner and Quan Vuong and Anna Walling and Haohuan Wang and Ury Zhilinsky},
      year={2024},
      howpublished={\url{https://arxiv.org/abs/2410.24164}}
}

@misc{shakey,
  title={Shakey the robot},
  author={Nilsson, Nils J},
  year={1984}
}

@INPROCEEDINGS{droiddataset, 
    AUTHOR    = {Alexander Khazatsky AND Karl Pertsch AND Suraj Nair AND Ashwin Balakrishna AND Sudeep Dasari AND Siddharth Karamcheti AND Soroush Nasiriany AND Mohan Kumar Srirama AND Lawrence Yunliang Chen AND Kirsty Ellis AND Peter David Fagan AND Joey Hejna AND Masha Itkina AND Marion Lepert AND Yecheng Jason Ma AND Patrick Tree Miller AND Jimmy Wu AND Suneel Belkhale AND Shivin Dass AND Huy Ha AND Arhan Jain AND Abraham Lee AND Youngwoon Lee AND Marius Memmel AND Sungjae Park AND Ilija Radosavovic AND Kaiyuan Wang AND Albert Zhan AND Kevin Black AND Cheng Chi AND Kyle Beltran Hatch AND Shan Lin AND Jingpei Lu AND Jean Mercat AND Abdul Rehman AND Pannag R Sanketi AND Archit Sharma AND Cody Simpson AND Quan Vuong AND Homer Rich Walke AND Blake Wulfe AND Ted Xiao AND Jonathan Heewon Yang AND Arefeh Yavary AND Tony Z. Zhao AND Christopher Agia AND Rohan Baijal AND Mateo Guaman Castro AND Daphne Chen AND Qiuyu Chen AND Trinity Chung AND Jaimyn Drake AND Ethan Paul Foster AND Jensen Gao AND David Antonio Herrera AND Minho Heo AND Kyle Hsu AND Jiaheng Hu AND Donovon Jackson AND Charlotte Le AND Yunshuang Li AND Roy Lin AND Zehan Ma AND Abhiram Maddukuri AND Suvir Mirchandani AND Daniel Morton AND Tony Nguyen AND Abigail O'Neill AND Rosario Scalise AND Derick Seale AND Victor Son AND Stephen Tian AND Emi Tran AND Andrew E. Wang AND Yilin Wu AND Annie Xie AND Jingyun Yang AND Patrick Yin AND Yunchu Zhang AND Osbert Bastani AND Glen Berseth AND Jeannette Bohg AND Ken Goldberg AND Abhinav Gupta AND Abhishek Gupta AND Dinesh Jayaraman AND Joseph J Lim AND Jitendra Malik AND Roberto Martín-Martín AND Subramanian Ramamoorthy AND Dorsa Sadigh AND Shuran Song AND Jiajun Wu AND Michael C. Yip AND Yuke Zhu AND Thomas Kollar AND Sergey Levine AND Chelsea Finn}, 
    TITLE     = {{DROID}: A Large-Scale In-The-Wild Robot Manipulation Dataset}, 
    BOOKTITLE = RSS, 
    YEAR      = {2024}
}

@inproceedings{robominddataset,
  author = {Wu, Kun and Hou, Chengkai and Liu, Jiaming and Che, Zhengping and Ju, Xiaozhu and Yang, Zhuqin and Li, Meng and Zhao, Yinuo and Xu, Zhiyuan and Yang, Guang and Fan, Shichao and Wang, Xinhua and Liao, Fei and Zhao, Zhen and Li, Guangyu and Jin, Zhao and Wang, Lecheng and Mao, Jilei and Liu, Ning and Ren, Pei and Zhang, Qiang and Lyu, Yaoxu and Liu, Mengzhen and Jingyang, He and Luo, Yulin and Gao, Zeyu and Li, Chenxuan and Gu, Chenyang and Fu, Yankai and Wu, Di and Wang, Xingyu and Chen, Sixiang and Wang, Zhenyu and An, Pengju and Qian, Siyuan and Zhang, Shanghang and Tang, Jian},
  booktitle= RSS, 
  title={{RoboMIND}: Benchmark on Multi-embodiment Intelligence Normative Data for Robot Manipulation}, 
  year={2025}
}

@article{ompl,
    Author = {Ioan A. {\c{S}}ucan and Mark Moll and Lydia E. Kavraki},
    Doi = {10.1109/MRA.2012.2205651},
    Journal = {{IEEE} Robotics \& Automation Magazine},
    Month = {December},
    Number = {4},
    Title = {The {O}pen {M}otion {P}lanning {L}ibrary},
    Volume = {19},
    Year = {2012}
}

@article{heo2023furniturebench,
  title={Furniturebench: Reproducible real-world benchmark for long-horizon complex manipulation},
  author={Heo, Minho and Lee, Youngwoon and Lee, Doohyun and Lim, Joseph J},
  journal=IJRR,
  year={2023},
  publisher={SAGE Publications Sage UK: London, England}
}

@inproceedings{simpler,
  title={Evaluating Real-World Robot Manipulation Policies in Simulation},
  author={Li, Xuanlin and Hsu, Kyle and Gu, Jiayuan and Mees, Oier and Pertsch, Karl and Walke, Homer Rich and Fu, Chuyuan and Lunawat, Ishikaa and Sieh, Isabel and Kirmani, Sean and others},
  year={2025},
  booktitle=CORL
}

@inproceedings{carpentier2019pinocchio,
   title={The Pinocchio C++ library -- A fast and flexible implementation of rigid body dynamics algorithms and their analytical derivatives},
   author={Carpentier, Justin and Saurel, Guilhem and Buondonno, Gabriele and Mirabel, Joseph and Lamiraux, Florent and Stasse, Olivier and Mansard, Nicolas},
   booktitle=SII,
   year={2019}
}

@article{sb3,
  author  = {Antonin Raffin and Ashley Hill and Adam Gleave and Anssi Kanervisto and Maximilian Ernestus and Noah Dormann},
  title   = {Stable-Baselines3: Reliable Reinforcement Learning Implementations},
  journal = jmlr,
  year    = {2021},
  volume  = {22},
  number  = {268}
}

@inproceedings{pbrs,
    title = {Policy Invariance Under Reward Transformations: Theory and Application to Reward Shaping},
    author = {Ng, Andrew Y. and Harada, Daishi and Russell, Stuart J.},
    year = {1999},
    booktitle = ICML,
}

\end{document}